%% file: main.tex
\documentclass[letterpaper]{article} % DO NOT CHANGE THIS
\usepackage{aaai25}  % DO NOT CHANGE THIS
\usepackage{times}  % DO NOT CHANGE THIS
\usepackage{helvet}  % DO NOT CHANGE THIS
\usepackage{courier}  % DO NOT CHANGE THIS
\usepackage[hyphens]{url}  % DO NOT CHANGE THIS
\usepackage{graphicx} % DO NOT CHANGE THIS
\usepackage{natbib}  % DO NOT CHANGE THIS AND DO NOT ADD ANY OPTIONS TO IT
\usepackage{caption} % DO NOT CHANGE THIS AND DO NOT ADD ANY OPTIONS TO IT
\usepackage{algorithm}
\usepackage{algorithmic}

\usepackage{newfloat}
\usepackage{listings}
\DeclareCaptionStyle{ruled}{labelfont=normalfont,labelsep=colon,strut=off} % DO NOT CHANGE THIS
\floatstyle{ruled}
\newfloat{listing}{tb}{lst}{}
\floatname{listing}{Listing}
\title{Compositionality Unlocks Deep Interpretable Models}
\author{
    Thomas Dooms\equalcontrib\textsuperscript{\rm 1}, Ward Gauderis\equalcontrib\textsuperscript{\rm 2},
    Geraint A.\ Wiggins\textsuperscript{\rm 2}, Jos\'e Oramas\textsuperscript{\rm 1}
}
\affiliations{
    \textsuperscript{\rm 1}Universiteit Antwerpen, Belgium\\
    \textsuperscript{\rm 2}Vrije Universiteit Brussel, Belgium\\
    thomas.dooms@uantwerpen.be, ward.gauderis@vub.be
}

\usepackage{booktabs}

\usepackage{tikz}
\usepackage{tikzit}
\input{book.tikzdefs}
\input{book.tikzstyles}

\newcommand{\rchi}{\raisebox{2pt}{$\chi$}}

\begin{document}

\maketitle

\begin{abstract}
We propose \raisebox{1.5pt}{$\chi$}-net, an intrinsically interpretable architecture combining the compositional multilinear structure of tensor networks with the expressivity and efficiency of deep neural networks. \raisebox{1.5pt}{$\chi$}-nets retain equal accuracy compared to their baseline counterparts. Our novel, efficient diagonalisation algorithm, ODT, reveals linear low-rank structure in a multilayer SVHN model. We leverage this toward formal weight-based interpretability and model compression.
\end{abstract}


\section{Introduction}

This paper presents preliminary results on developing interpretable neural networks while retaining competitive accuracy. We introduce the $\rchi$-net architecture and the ODT algorithm, which diagonalises and truncates such networks. We discuss the background and rationale behind these networks and provide evidence that these networks are easily interpretable based on their weights. The contributions of this work are as follows:

\begin{itemize}
    \item We propose $\rchi$-net, an intrinsically interpretable architecture with competitive accuracy.
    \item We propose the ODT algorithm to diagonalise the $\rchi$-net and truncate it up to arbitrary precision.
    \item We extract meaningful learned patterns from the resulting highly structured model.
\end{itemize}

\subsubsection{Mechanistic interpretability} provides a microscope into a neural network's inner mechanisms, without which we cannot meaningfully verify their safety, reliability, or alignment with human values. The field aims to reverse-engineer deep neural networks into a set of understandable algorithms or components. Just as one decompiles a computer program into variables and functions, researchers aim to reduce deep networks into a collection of semantically meaningful circuits \cite{intro_circuits}. Such circuits are broadly defined as small or sparse `computation' graphs \citep{intro_circuits, loconteWhatRelationshipTensor2024}.
However, due to the inherent non-decomposability of many modules in neural networks, this task is often rooted in post-hoc approximations \citep{monosemanticity}. 
Detrimentally, since interpretability is an ill-defined metric \citep{falsifiable_interpretable}, much research cherry-picks examples and draws incomplete conclusions, following a practice referred to as \textit{streetlight interpretability} \citep{streetlight}. Interpretability should focus on finding mechanisms important to the model, not humans. Our results indicate that rooting interpretability in rigorous decompositions is a promising avenue to achieving this.

\subsubsection{Compositional AI} emphasises understanding an artificial system's behaviour by examining how its constituent processes interact and compose, rather than studying components in isolation \citep{coeckePicturingQuantumProcesses2018,coeckeCompositionalityWeSee2021}. It uses the principle of {\it compositionality} from categorical quantum mechanics -- studying systems through their components and interactions -- to provide a common mathematical foundation for symbolic and subsymbolic AI, and to break the curse of dimensionality in learning and understanding complex models.
The field of \textit{Applied category theory in AI} seeks to provide an overarching framework that unifies theory and practice \citep{shieblerCategoryTheoryMachine2021,dudzikCategoriesAI2022} through its ability to structure, uncover and facilitate cross-disciplinary insights.
By formalising the compositional structure, neuro-symbolic methods can integrate symbolic domain knowledge to improve data efficiency and interpretability of subsymbolic models. \citep{lorenzQNLPPracticeRunning2023}.
% By formalising and Compositional models can serve as neuro-symbolic methods that impose symbolic structure in model design while facilitating the interpretation of subsymbolic models.
% Formalising and understanding the compositional structure of models and their domains can both improve data-efficiency by integrating domain knowledge and interpretability \citep{lorenzQNLPPracticeRunning2023}.
Categorical deep learning uses compositional domain structure to derive inductive biases in deep learning 
\citep{gavranovicCategoricalDeepLearning2024,dudzikCategoriesAI2022,bronsteinGeometricDeepLearning2021}. Categorical string diagrams \citep{coeckePicturingQuantumProcesses2018} formalise \textit{compositional models} in symmetric monoidal categories and recognise `intrinsically interpretable' models as \textit{compositionally-interpretable models} \citep{tullCompositionalInterpretabilityXAI2024}.

% language
\subsubsection{Rationale}
String diagrams interpreted in the category of real finite-dimensional Hilbert spaces create tensor network diagrams where boxes represent linear maps, and wires represent vector spaces \citep{coeckePicturingQuantumProcesses2018}.
Tensor networks offer rich compositional structures that enable principled network design \citep{levineQuantumEntanglementDeep2018} and can serve as representations bridging symbolic and subsymbolic reasoning methods \citep{gauderisQuantumTheoryKnowledge2023}.
% neural vs tensor
Tensor networks capture structural relations in high-dimensional Hilbert spaces, but are limited to linear relations. Neural networks address this with non-linear activation functions, yet the composition of non-linearities remains poorly understood.
% linear
This work bridges the representational power of non-linear neural networks and the compositional structure of linear tensor networks, which has been shown capable of retaining near-SoTA accuracy while providing appealing mathematical properties (\citet{bilinear, chengMULTILINEAROPERATORNETWORKS2024}; inter alia).
% \vspace{-.5em}

\section{$\rchi$-net Architecture}
We propose the deep $\rchi$-net (CHI-net) architecture,
short for \textit{Compositionally and Hierarchically Interpretable network}.
Through appropriate coordination of tensor composition, principled weight-tying
patterns, and a non-linear cloning operator, $\rchi$-nets can be efficiently implemented, trained, and evaluated
as a deep non-linear neural architecture, but analysed, recomposed and interpreted as a
linear tree tensor network, combining the best of both worlds.

% Polynomial -> multilinear -> linear
Using a non-linear cloning operator, multi-variable polynomials are learned as multilinear maps that can be reinterpreted as linear maps over a tensor product space.
% 1 non-linear cloning operator: polynomial as multilinear
Inspired by the linear no-cloning theorem \citep{coeckePicturingQuantumProcesses2018}, we introduce the non-linear cloning operator, a purely diagrammatic manipulation:
\etz{cloning}{0.8}
% The cloning operator works both as shorthand notation and as an indicator of the computationally efficient direction of the contraction.
This unfolds into a multilinear map to which multiple copies of the input are fed.
Cloning has already been leveraged implicitly in network design and interpretability \citep{bilinear, chengMULTILINEAROPERATORNETWORKS2024, dubeyScalableInterpretabilityPolynomials2022}.
% 2 tensor (de)composition to avoid exponential growth of parameters
Through (de)composition, tensor networks avoid exponential parameter growth when learning higher-order multilinear maps \citep{chrysosDeepPolynomialNeural2020,grelierLearningTreebasedTensor2018}.
% 3 linear
Since any multi-linear map can be represented as a linear map over a tensor product space, the network can be viewed as a linear map with weight-tying induced by the cloning operator over a high-dimensional kernel vector space.
%\vspace{-.3em}

% \subsection{$\rchi$-net architecture}
The architecture of a $\rchi$-net $\rchi: \mathcal{I} \to \mathcal{O}$ from input space
$\mathcal{I}$ to output space $\mathcal{O}$ is shown in Figure~\ref{fig:chinet_embedding}.
The input space $\mathcal{I}$ is embedded into a \textit{hidden bond space}
$\mathcal{H}_1$ by a linear
embedding map $e: \mathcal{I} \to \mathcal{H}_1$. This feeds into $L$ layers of the cloning
operator followed by a third-order tensor \textit{core} $f_i: \mathcal{H}_i \to
\mathcal{H}_{i+1}$ for
$i = 1, 2, \ldots, L$. The output space $\mathcal{O}$ is obtained by unembedding
$\mathcal{H}_{L+1}$ using a linear unembedding map $u: \mathcal{H}_{L+1} \to \mathcal{O}$.
For notational simplicity, the maps $e$ and $u$ are also
referred to as $f_0$ and $f_{L+1}$, respectively.
When the output $\mathcal{O}$ is a probability distribution, the unembedding map $u$ can be
fed into a softmax function, omitted here for simplicity.

Intuitively, a $\rchi$-net replaces the standard hidden layer consisting of a linear map and a non-linear activation function by a cloning operator and a multilinear map.
The network's forward pass is
efficiently computed bottom-up, starting from the input $x \in \mathcal{I}$ and evaluating the
cores $f_i$, in sequence, alternating with the cloning operator, as shown in Figure~\ref{fig:chinet_embedding}. 
By unrolling the cloning operators using Equation~\ref{eq:cloning}, the network can be interpreted
as a tree tensor network featuring weight-tying patterns per layer.

\ctz{chinet_embedding}{0.73}{
		String diagram of a 3-layer $\rchi$-net with input $x \in \mathcal{I}$,
		linear embedding and unembedding maps $e: \mathcal{I} \to \mathcal{H}_1$ and $u:
		\mathcal{H}_4 \to \mathcal{O}$,
		and multilinear cores $f_i: \mathcal{H}_{i} \to \mathcal{H}_{i+1}$ for $i = 1, 2, 3$.
		On the left side, the network is interpreted bottom-up, reflecting
		the efficient forward pass evaluation.
		The right side contains the unfolded tree tensor network, where
		contraction is no longer
		restricted to unidirectional evaluation. The expansion introduces weight-tying
		patterns per layer.
}

Every core $f_i$ is a bilinear map of its inputs $\mathcal{H}_i \otimes \mathcal{H}_i$.
Combined with the non-linear immediately preceding cloning operator, $f_i$ captures quadratic
interactions between inputs. 
A $\rchi$-net of depth $L$ can only express multilinear polynomials over its inputs with monomials of degree $2^L$. This restriction to high-degree interactions is detrimental to generalisation, as real-world problems often depend on lower-degree terms \citep{linWhyDoesDeep2017,liTaylorTheoremNew2021}. Adding a constant input as a bias term overcomes this limitation: 
$x \leftarrow (1, x)$.
With this modification, an $L$-layer $\rchi$-net can express multilinear polynomials of degree up to $2^L$.
We impose further structure on the cores to restrict the number of learnable parameters. Specifically, each core $f_i$ is parametrised using a transposed Khatri-Rao product of two matrices $A_i$ and $B_i$:
\etz{bilinear}{0.8}
Here, the white circle represents the third-order tensor identified by a Kronecker delta
$\delta_{jk}^l$ over
input indices $j$ and $k$ and output index $l$ and coincides with an element-wise
multiplication of the input vectors \citep{coeckePicturingQuantumProcesses2018}.
This parameterization, known as a \textit{bilinear layer} \citep{glu_variants, technical_note_bilinear}, is increasingly popular in neural network design but is less expressive than a full bilinear map.
Due to the preceding cloning operator, the bilinear layer's parameterisation can be symmetrised, with respect to its inputs, post-training without affecting its expressivity:
%by averaging the two permutations of the input indices of the contracted core tensor:
\[(f_i)_{jk}^{l} \leftarrow \frac{1}{2} \left( (f_i)_{jk}^{l} + (f_i)_{kj}^{l} \right),
\quad \forall i, j, k\]
Throughout, we assume this symmetry is enforced.
% \vspace{-.5em }

\section{ODT Algorithm}  \label{sec:ODT}
We present the novel \textit{Orthogonalisation, Diagonalisation, and Truncation} (ODT) algorithm,
inspired by the hierarchical singular value decomposition (HSVD) for tree-structured
tensor networks \citep{oseledetsBreakingCurseDimensionality2009a,grasedyckHierarchicalSingularValue2010}. The ODT algorithm efficiently compresses trained $\rchi$-nets and extracts their low-rank interpretable features.
The overall time complexity of the algorithm is $\mathcal{O}(L \cdot h^4)$, with hidden dimension $h$ and depth $L$, discussed in Appendix~\ref{app:complexity}.
Each step is discussed individually, and a compression error bound is given. 
Accompanying diagrammatic proofs and further explanations are presented in Appendix~\ref{app:diagrams}.
% \vspace{-.75em}

\subsubsection{Orthogonalisation}
% While the first step is not strictly necessary for model truncation, it greatly reduces the
% number of tensor contractions required in the subsequent steps, and was found to improve
% the interpretability of the resulting cores by making them sparse and low-rank.
The objective is to make the cores $f_i$ orthonormal for $i=0, \dots, L$ without changing the network behaviour.
This is done bottom-up, starting from the input embedding $e = f_0$ and proceeding to the final
core $f_L$, by applying the RQ decomposition to each matricised core $f_i$, where $f_i$ is replaced
by the unitary $Q_i$ and contracting the non-orthogonal part $R_i$ into the next core $f_{i+1}$.
To ensure that the bond dimensions do not grow during this process, reduced RQ decomposition is
used so that the adjoints of the new cores $f_i^\bot = Q_i$ are isometric with orthonormal columns.
% An example of the orthogonalisation process is shown in Figure~\ref{fig:orthogonalisation}.
The result is a network in which all cores $f_i$ are isometries, 
% which means they can be interpreted as discarding non-informative interactions
except for the unembedding $u = f_{L+1}$, which contains the non-orthogonal part of the network.
% The isometric property of the orthogonalised $\rchi$-net is shown in Figure~\ref{fig:isometric}.
% \vspace{-1em}

\subsubsection{Diagonalisation}
This step computes an orthogonal projection matrix $P_i$ for each
bond $H_i$ that projects onto the left singular vectors of the $\rchi$-net,
matricised with respect to a single $H_i$ bond. These projectors are computed by diagonalising the
Gram matrix $G_i$ of the $\rchi$-net for that bond.
This calculation, illustrated in Figure~\ref{fig:svd}, 
consists of three steps. First, the $\rchi$-net is composed with its adjoint and the tensor
contraction at a bond $H_i$ is removed. This results in a self-adjoint Gram matrix $G_i: H_i \to H_i$.
Second, its spectral decomposition $G_i = V_i \Lambda_i V_i^*$ is computed.
$V_i$ is the unitary that diagonalises $G_i$ and contains the left-singular vectors of
the matricised $\rchi$-net,
and $\Lambda_i$ is a diagonal matrix of the corresponding real and positive singular values,
sorted in decreasing order of magnitude.
Finally, the self-adjoint, idempotent and therefore orthogonal projector $P_i$ on the
left singular vectors are constructed as:
\begin{equation}
    P_i = V_i V_i^* 
    \label{eq:projector} 
\end{equation}
% \etz{projector}{0.8}
%By \citet[Proposition 5.77]{coeckePicturingQuantumProcesses2018},
Inserting $P_i$ into bond $H_i$ of the
$\rchi$-net does not affect its behaviour \citep[Prop.\ 5.77]{coeckePicturingQuantumProcesses2018}.
% shown on the left of Figure~\ref{fig:parallel}.
%\vspace*{-.3em}

\subsubsection{Truncation}
Given the singular vectors and values of each bond $H_i$, the last step is to reduce
the rank $r_i$ of each projection $P_i$ to project only onto the $r_i$ most important
singular vectors.
This is achieved by truncating the unitary $V_i$ to the first $r_i$ columns (preserving
isometric properties) so that
$\text{dim}(\mathcal{H}'_i) = r_i$ in Equation~\ref{eq:projector}, while the truncated projector
is the low-rank approximation that minimises the Frobenius distance to the original projector.
Finally, the truncated projectors $P_i$ are partially contracted in both $f^\bot_{i-1}$
and $f^\bot_i$.
% as shown in Figure~\ref{fig:truncation}. 
In this way,
the new bond dimensions of $\mathcal{H}'_i$ are reduced to
$r_i$ in the truncated $\rchi'$-net.
% as shown on the right side of Figure~\ref{fig:parallel}.
Note that the truncated $\rchi'$-net is a low-rank approximation of the original $\rchi$-net that
retains the symmetric isometric properties imposed by the symmetrisation and
orthogonalisation steps.

We establish the compression error bounds by following the proof for HSVD
\citep{grasedyckHierarchicalSingularValue2010,grelierLearningTreebasedTensor2018}. Let $\epsilon > 0$ be any threshold. If each $G'_i$ is the truncated $G_i$ with rank $r_i$ such that:
\[\| G_i \|^2_F - \| G'_i \|^2_F \leq \frac{\epsilon^2}{2^{(L + 1)}-1} \| G_i \|^2_F\]
where the denominator is the number of projections in the unfolded tensor network, then the
truncated $\rchi'$-net is a low-rank approximation of the original $\rchi$-net with
relative precision $\epsilon$:
\[\| \rchi - \rchi' \|_F \leq \epsilon \| \rchi \|_F\]
The prospect of bounding the loss is left for future work.
% \vspace{-1em}

\section{Experiments} \label{sec:experiments}

\begin{figure}
    \centering
    \includegraphics[width=\linewidth]{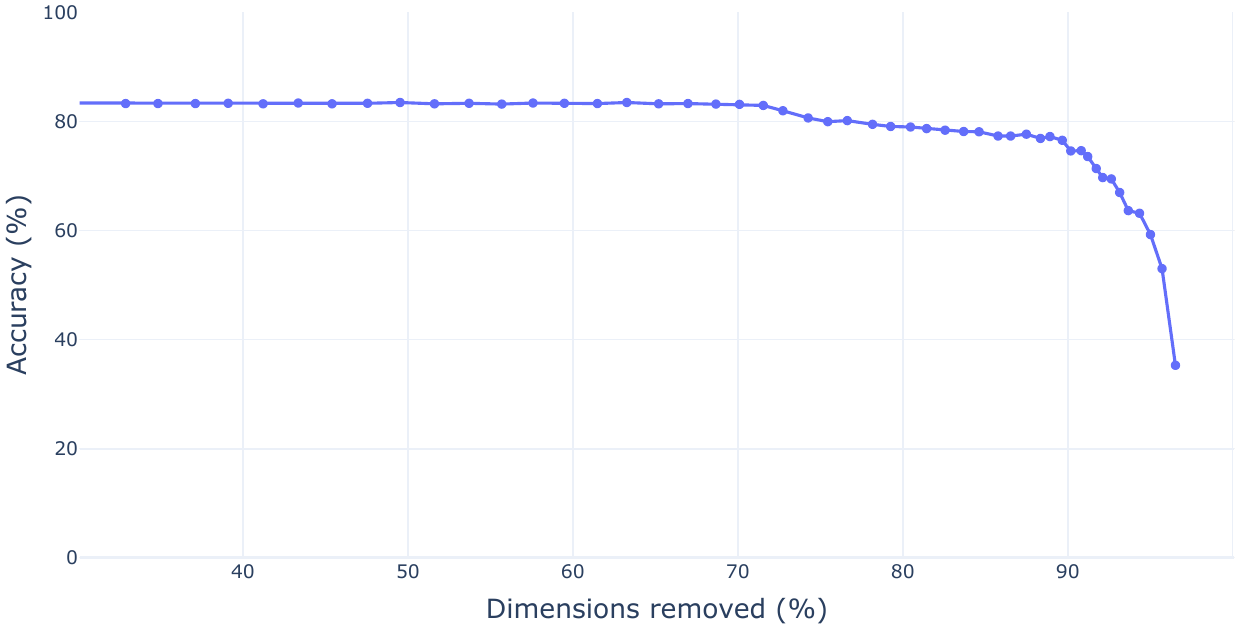}
    \caption{The accuracy curve shows that it is possible to truncate about 70\% of the model's dimensions without compromising accuracy. About 90\% of dimensions can be truncated with a small drop in accuracy.}
    \label{fig:trunc_acc}
\end{figure}

We consider a 3-layer model trained on the SVHN dataset (a real-world variant of MNIST). This model achieves about 85\% test accuracy. Details about the setup can be found in Appendix~\ref{app:setup}. The diagonalised cores are sparse, and the truncation discards about 70\% of bond dimensions with no decrease in accuracy (Figure~\ref{fig:trunc_acc}). This section discusses leveraging this structure to extract learned patterns from the model.

\subsubsection{Weight interpretation}
The $\rchi$-net consists of an embedding matrix ($e=f_0$), whose rows are called \textit{atoms}, and core tensors ($f_1$-$f_3$), whose symmetric input slices are called \textit{interaction matrices}. We call a composition of atoms a \textit{feature} and define its \textit{activation} as its composition with an input. An atom's interpretation depends on its subsequent interaction matrix; we distinguish between three kinds. Diagonal entries show an input's self-multiplication, with the weight determining positive or negative value. Off-diagonal entries act as gates by multiplying different inputs together. Entries on the first row/column are scalar values since they multiply with a constant (the bias). All figures show positive weights in blue and negative weights in red.

\subsubsection{Tensor anatomy}
\begin{figure}
    \centering
    \includegraphics[width=\linewidth]{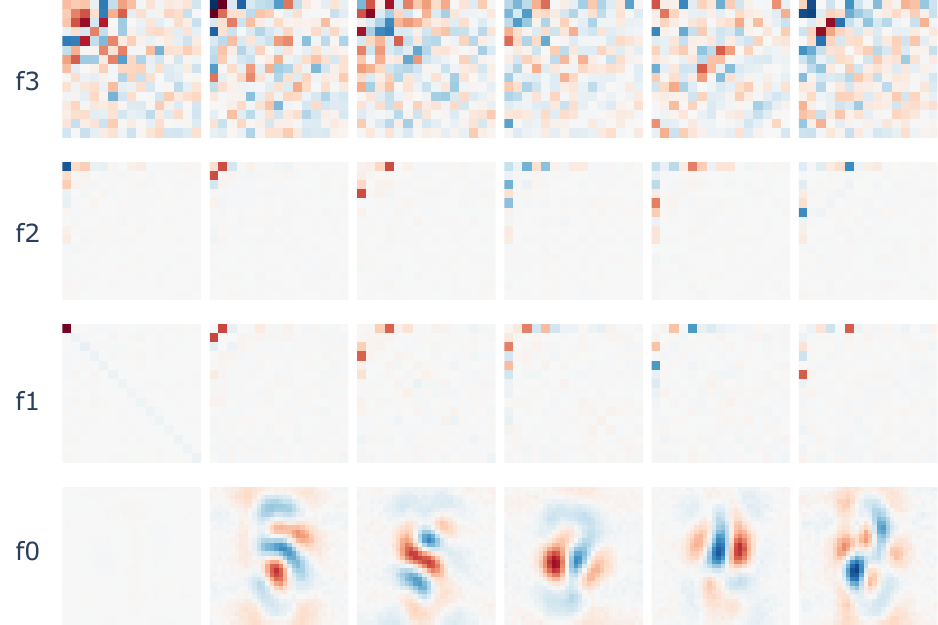}
    \caption{The six most important atoms ($e \mathord{=} f_0$) and interaction matrices ($f_1$-$f_3$) for each layer of the decomposed model. The unembedding ($u \mathord{=} f_4$) is contracted into $f_3$ for brevity. The atoms are reshaped into the image's dimensions for visual clarity; they contain patterns such as edge detectors and proto-digits. The middle interaction matrices are highly sparse and are dominated by constant interactions. The root ($f_3$) is denser, combining many learned features.}
    \label{fig:cores}
\end{figure}

Our diagonalised model's atoms are interpretable, and its interaction matrices (except the root) are sparse, as shown in Figure~\ref{fig:cores}. Interestingly, the constant dimension stays separated despite not being hard-coded into the diagonalisation; it does not match with any input except the appended bias (not shown in the figure). The other atoms have coherent spatial structures; some resemble proto-digits, while others are localised edge detectors. Constant interactions dominate the interaction matrices of $f_1$ and $f_2$, comprising about 90\% of their norm. This effectively corresponds to sparsely summing certain atoms; note that this is a fully linear operation. Local decompositions miss this structure, as discussed in Appendix~\ref{app:ablations}. The root ($f_3$) is denser, capturing more intricate interactions. Its effective input dimension is about 13, indicating that it is still structured.

\subsubsection{Digit extraction}
We extract interpretable features from this model corresponding to the output classes. Our approach is inspired by \citet{bilinear}, who consider the highest eigenvector for a given output class and show them to be interpretable in single-layer models. 
Our models span multiple layers, so we cannot simply reuse this approach. 
However, given that linear interactions dominate the first few layers, it is possible to directly use this to project the vectors onto the input space. This approach yields highly-interpretable digits (Figure~\ref{fig:digits}). The other eigenvectors of digits are covered in Appendix~\ref{app:features}. In short, using the near-linearity found by the decomposition provides an excellent lens through which to understand latent features.

\begin{figure}
    \centering
    \includegraphics[width=\linewidth]{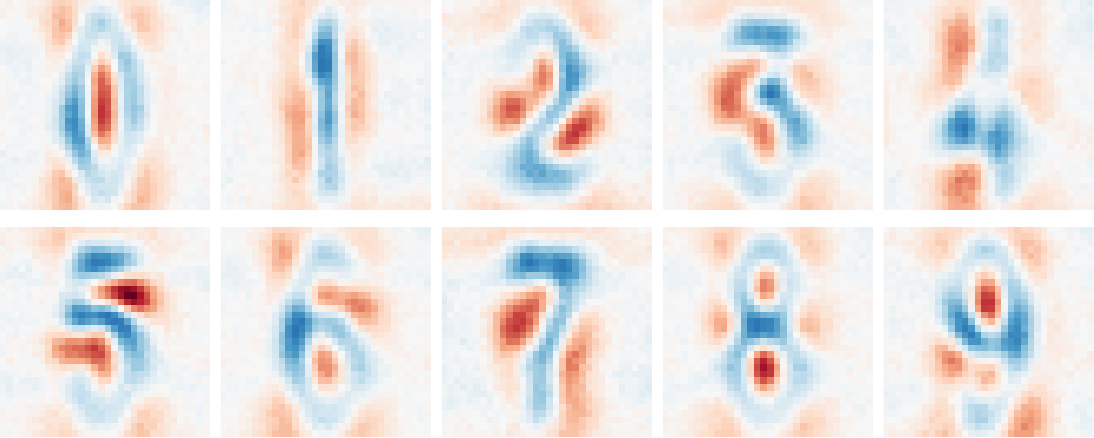}
    \caption{The most important eigenvector of the root core ($u \circ f_3$) per digit. These are linearly traced through the previous cores onto the input space. These represent the prototypical digits from the training data. }
    \label{fig:digits}
\end{figure}

\section{Conclusion}
\subsubsection{Summary} This paper introduces $\rchi$-nets, an interpretable neural network architecture that maintains competitive accuracy. We present the ODT algorithm, which efficiently diagonalises $\rchi$-nets, revealing interpretable low-rank linear structures in multilayer models and allowing model compression by ranking hidden dimensions. Through tensor composition, weight-tying, and a non-linear cloning operator, $\rchi$-nets can be implemented as deep neural networks but analysed as tree tensor networks, combining the advantages of both.

% \subsubsection{Implications} Many interpretability works rely on local "bag-of-features" approaches, where it is hard to compute a feature's importance and downstream effects. By embracing the compositional structure of neural networks, this paper offers a significant step toward solving this.

\subsubsection{Ongoing work}
Despite the scalability of the proposed architecture and algorithm, this paper focuses on small models. While the ODT algorithm identifies shallow neural circuits, deeper models may involve complex multilayer interactions, which will be harder to interpret comprehensively. Ongoing work includes extending this approach to modern architectures, exploring probabilistic generative modelling \citep{loconteWhatRelationshipTensor2024}, and involving the dataset in the decomposition to capture input statistics or integrate metadata.

% cloning operator as explicit kernel map (so we don't need to explain it in this paper)

% While our proposed networks have been shown to scale to complex tasks using modern architectures, this paper focuses on small models. Our current interpretability methods rely on the model's near-linear behaviour; we expect more challenging datasets to have more complex interactions spanning multiple layers, making linearisation less beneficial.  
% However, there is reason to believe many circuits in deep networks are inherently shallow \citep{shallow_ensembles}, which this technique can uncover quite well. Furthermore, we have found intermediate features (which consider the network up to a given layer) are quite interpretable.

%%%%%%%%%%%%%%%%%%%%%%%%%%%%%%%%%%%%%%%%%%%%%%%%%%%%%%%%%%%%%%%%%%%%%%%%%%%%ù

% \clearpage
\appendix
\section*{Appendix}

\section{Experimental setup} \label{app:setup}
% This section provides additional details about the model used throughout experiments.

\subsubsection{Dataset} To train our model, we use the SVHN dataset, which is essentially a harder version of MNIST. The motivation is that this provides a reasonable middle ground between MNIST, which is solvable using linear models, and ImageNet, which requires huge models. Furthermore, SVHN has easily discernible classes, making it a perfect testbed for our purposes. We concatenate the "train" and "extra" splits to train and use the "test" split to measure accuracy. Our input processing step grayscales the images and adds Gaussian noise; we use no other regularization.

\subsubsection{Architecture} The discussed model consists of 3 hidden (bilinear) layers surrounded by an embedding and unembedding (head). These hidden layers all use biases of the form. The embedding projects the input onto a 256-dimensional latent space, and the unembedding projects this onto 10 the output classes. This is followed by a softmax operation to acquire the prediction probabilities. We use a variant of BatchNorm as normalisation that only divides by the $L_2$ norm (akin to RMSNorm). After training, this single average is contracted into its neighbouring matrix.

\begin{table}[H]
    \centering
    \begin{tabular}{lr}
        \multicolumn{2}{c}{\textbf{Hyperparameters}} \\
        \midrule
        \textbf{input noise norm} & 0.3 \\
        \textbf{weight decay} & 1.0 \\
        \textbf{learning rate} & 0.001 \\
        \textbf{normalisation} & RmsBatchNorm \\
        \textbf{batch size} & 2048 \\
        \textbf{optimiser} & AdamW \\
        \textbf{schedule} & cosine \\
        \textbf{epochs} & 20 \\
    \end{tabular}
    \label{tab:hyperparameters}
    \caption{Training setup for the studied model(s).}
\end{table}

\section{$\rchi$-net ablations}
This work aims to establish deep $\rchi$-nets as a viable and interpretable alternative to ordinary multi-layer perceptrons. Previous works \citep{bilinear, glu_variants, chengMULTILINEAROPERATORNETWORKS2024} provide evidence that multilinear networks can retain competitive accuracy when scaling both dataset complexity and model size. Specifically, \citet{bilinear} finds transformers that use bilinear layers to be 6\% less sample efficient. In a data-abundant regime, training for slightly longer can offset this. However, in a data-constrained regime, training for more epochs may not close that gap. Empirically, we find that ReLU-based networks benefit from training for more epochs while $\rchi$-nets sometimes do not. 

We perform a set of ablations on our small models to show that $\rchi$-nets achieve near-accuracy parity with their baseline counterparts (Table~\ref{tab:accuracy}). These experiments follow the setup from Appendix~\ref{app:setup}, only varying the depth. 

\renewcommand{\arraystretch}{1.2}
\begin{table}
\centering
\begin{tabular}{l|c|c|c|c}
                & \textbf{1 layer} & \textbf{2 layers} & \textbf{3 layers} & \textbf{4 layers} \\
\hline
\textbf{$\rchi$-net} & 83.8\% & 84.2\% & 85.4\% & 86.5\%  \\
\textbf{Baseline}    & 85.7\% & 86.4\% & 87.3\% & 88.0\%  \\
\end{tabular}
\caption{Accuracy of $\rchi$-nets and a ReLU baseline across depths. A linear model only attains 23\%.}
\label{tab:accuracy}
\end{table}
\renewcommand{\arraystretch}{1.0}

% \begin{table}[]
% \begin{tabular}{l|c|c|c|c}
%             & \textbf{1 layer} & \textbf{2 layers} & \textbf{3 layers} & \textbf{4 layers} \\
% \hline
% \textbf{$\rchi$-net} & 72.9\%  & 72.9\%   & 72.8\%   & 73.7\%  \\
% \textbf{Baseline}    & 70.9\%  & 71.0\%   & 70.3\%   & 72.5\%  \\
% \end{tabular}
% \end{table}
% \renewcommand{\arraystretch}{1.0}

% \subsubsection{Dimensions} We show the effective dimensionality, which reveals that the truncation does not impact this metric until the last few dimensions. Importantly, we compute the effective dimensions without the bias dimension, which heavily skews this metric. When including the biases, the effective dimensions of bonds 1 and 2 are slightly above 2, indicating the layer to be almost fully linear.

% \begin{figure}
%     \centering
%     \includegraphics[width=\linewidth]{figures/trunc_effective.pdf}
%     \caption{The effective dimension remains stable when removing dimensions only that fall sharply near the end. This decline occurs simultaneously with the drastic change in loss and accuracy. }
%     \label{fig:trunc_effective}
% \end{figure}

\section{Explaining predictions}
We employ our extracted features to evaluate their usefulness in explaining a given classification output. As building blocks in our explanation, we use the eigenvectors for all digits, of which the highest ones are shown in Figure~\ref{fig:digits} and Figure~\ref{fig:features}. We then measure the activation of each feature, which is computed as their inner product with the input. These activations are then multiplied by the eigenvalues for that feature.

\begin{figure}
    \centering
    \includegraphics[width=\linewidth]{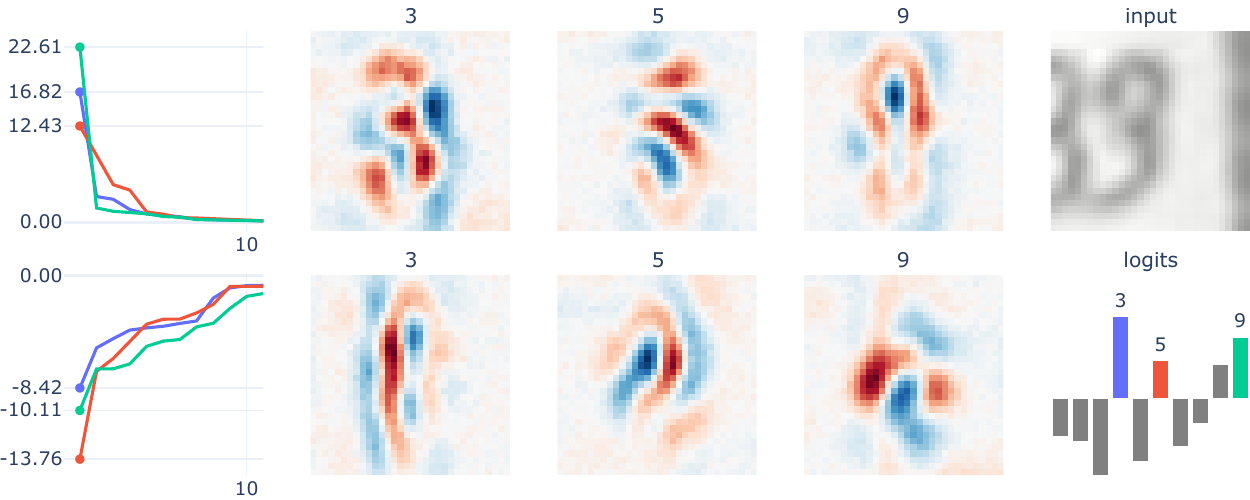}
    \caption{Using extracted features from the model to explain the classification logits. The left shows the importance scores for all features (split by positive and negative contribution), the most important ones of which are shown in the middle section. The right shows the logits along with the evaluated input.}
    \label{fig:explain}
\end{figure}

In Figure~\ref{fig:explain}, we consider a particular 3 almost resembling a 9 (possibly due to the grayscaling). This shows a feature for the digit 9 has the highest positive impact. However, a long tail of negative features inhibits the digit 9, resulting in a slightly higher logit for the digit 3.

\section{Additional features} \label{app:features}
\begin{figure}[!t]
    \centering
    \includegraphics[width=\linewidth]{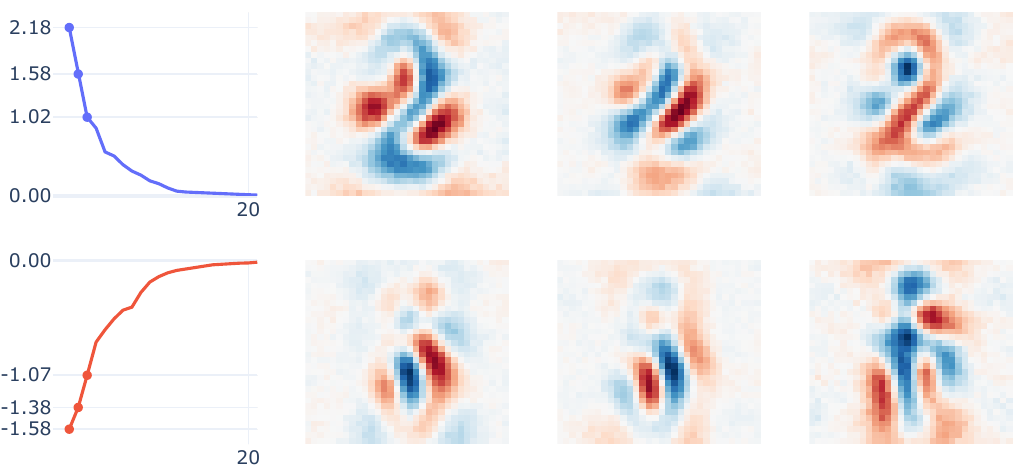}
    
    \vspace{2em}
    
    \includegraphics[width=\linewidth]{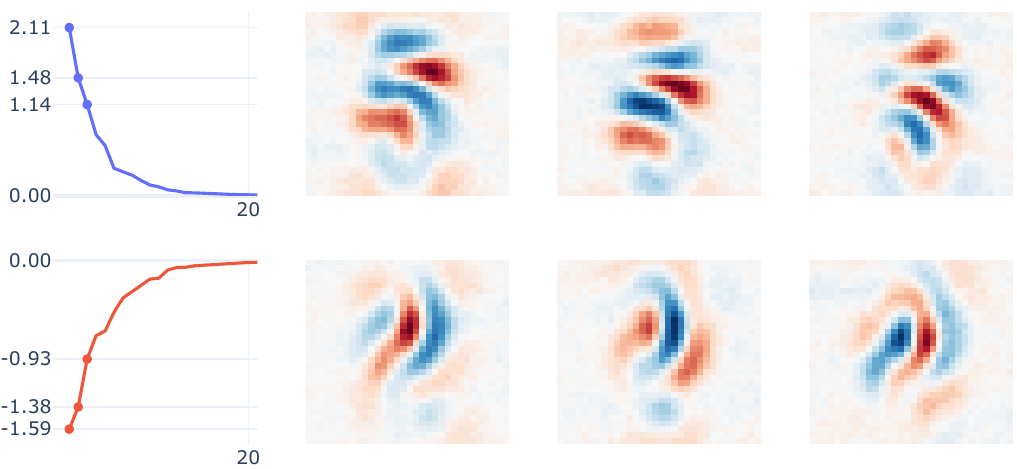}
    
    \vspace{2em}
    
    \includegraphics[width=\linewidth]{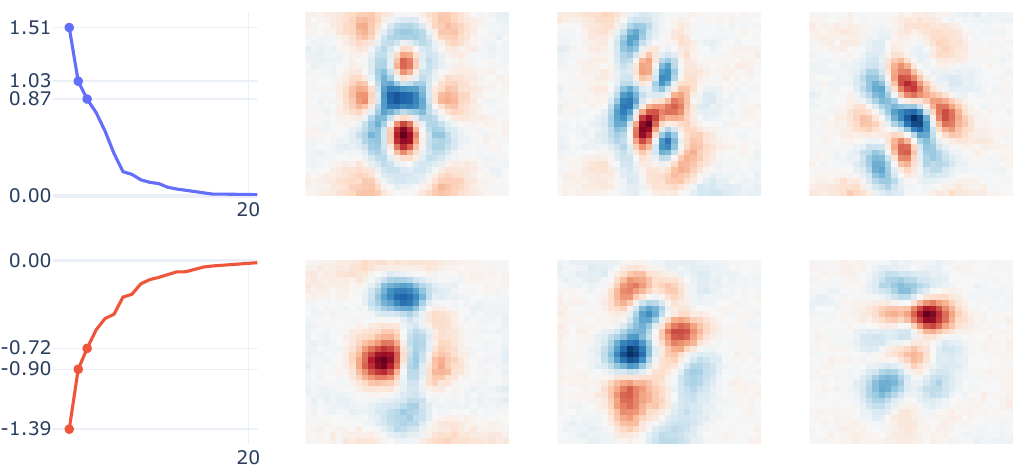}
    \caption{Most important projected eigenvectors for the digits 2 (top), 5 (middle) and 8 (bottom). The left graphs indicate the magnitude of the eigenvalues, split by positive and negative. The right side depicts the eigenvectors corresponding to the indicated eigenvalues. }
    \label{fig:features}
\end{figure}

Each digit has multiple eigenvectors; this section discusses the top eigenvectors and the eigenvalues for the digits 2, 5 and 8 (shown in Figure~\ref{fig:features}). The first positive eigenvector often indicates a prototypical instance of the given digit, the second the most salient stroke, and the third depends on the digit. 
Strokes are often detected by Gabor-like filters, matching across a range of possible locations.

The negative eigenvectors are slightly harder to interpret but correspond to a stroke that would strongly change the given digit to another. For instance, the negative eigenvectors of a 5 correspond to a stroke closing the top part; if that were present, the feature would likely become a 9.

% \section{Anecdotal observations}
% This section contains a collection of interesting observations, which we do not make precise and should hence not be seen as fact. However, we still think these are worth sharing.

% \subsubsection{Dataset size has a drastic effect on interpretability.} Even though MNIST is a simpler dataset, we found that training on SVHN leads to more interpretable models. Informally, it seems (to some degree) that a model can simply memorise MNIST digits instead of learning generalizing patterns. Relatedly, it has been observed that larger models admit more internal structure \citep{circular_features}.

% \subsubsection{Many epochs result in less interpretable features.}
% Even though we use input noise regularization, models tend to learn more spurious concepts after an extended number of epochs. This may be related to the findings of \citet{statistics_increasing_complexity}.

\section{ODT in numbers} \label{app:ablations}
While the ODT algorithm has a solid theoretical foundation, this section discusses how that translates to practice. We compare ODT to SVD and perform numerous ablations concerning the truncation procedure.

\subsubsection{Rank} 
\begin{figure}
    \centering
    \includegraphics[width=\linewidth]{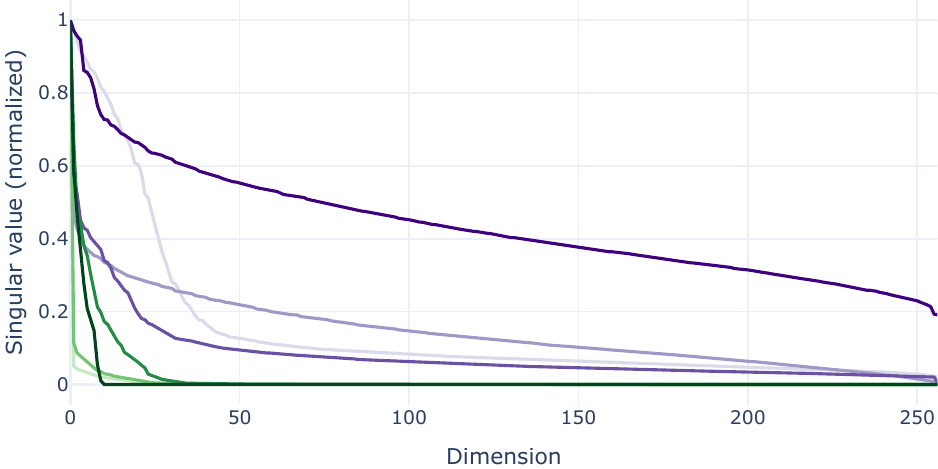}
    \caption{The normalized singular values of the bond dimension (increasing opacity with depth) for SVD (purple) and ODT (green). SVD does not find a clean low-rank structure, while ODT does.}
    \label{fig:local}
\end{figure}

We compare the ODT algorithm versus the commonly used SVD to measure the bond dimensions of our model (Figure~\ref{fig:local}). SVD has a very long tail, indicating none of the core tensors are low-rank by themselves. In contrast, the singular values from ODT immediately drop to zero; it is only by diagonalising that their structured nature becomes apparent. The effective dimensions are shown in Table~\ref{tab:effective}.

An effective bond dimension smaller than 2 indicates the model is fully linear (only the constant dimension is used). This is almost the case in the first two layers. Computing the effective dimension without the constant results in about 14 for all layers. We believe this to reflect the dataset's intrinsic dimensionality.

\renewcommand{\arraystretch}{1.2}
\begin{table}
\centering
\begin{tabular}{l|cccc}
    & \textbf{Bond 0} & \textbf{Bond 1} & \textbf{Bond 2} & \textbf{Bond 3} \\
\hline
\textbf{SVD} & 83.4   & 165.9  & 100.5  & 229.2  \\
\textbf{ODT} & 2.1    & 3.3    & 13.3   & 5.7   
\end{tabular}
\caption{Comparison of the effective ($L_2$) bond dimensions for SVD and ODT.  }
\label{tab:effective}
\end{table}

\subsubsection{Loss} 
\begin{figure}
    \centering
    \includegraphics[width=\linewidth]{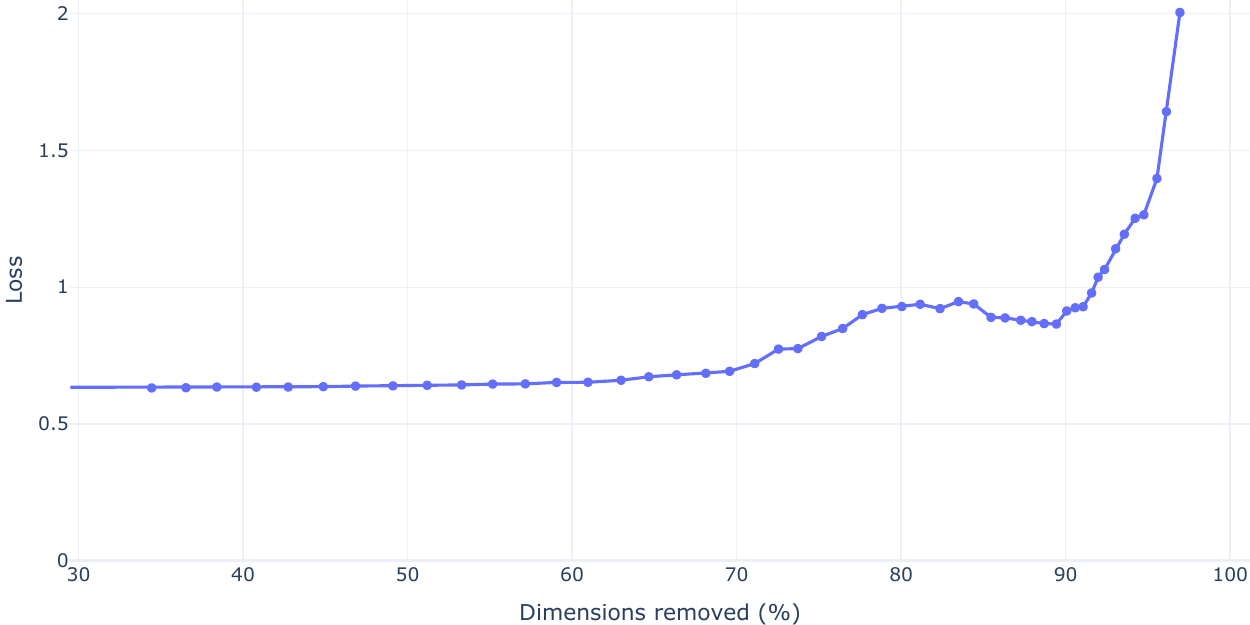}
    \caption{The loss in function of removed dimensions; large parts can be removed without affecting loss.}
    \label{fig:trunc_loss}
\end{figure}

In addition to Figure~\ref{fig:trunc_acc}, we provide the resulting loss after any truncation (Figure~\ref{fig:trunc_loss}). This paints a similar picture as the accuracy comparison, but there is a dip in the loss after truncating about 90\% of dimensions. We are currently uncertain about its cause. One hypothesis is that the network has a generalizing backbone and a long tail of spurious correlations. Removing a part of these correlations may destabilise the model, but removing all of them is beneficial. Once parts of the backbone are removed, the loss spikes.

\subsubsection{Norm} 
\begin{figure}
    \centering
    \includegraphics[width=\linewidth]{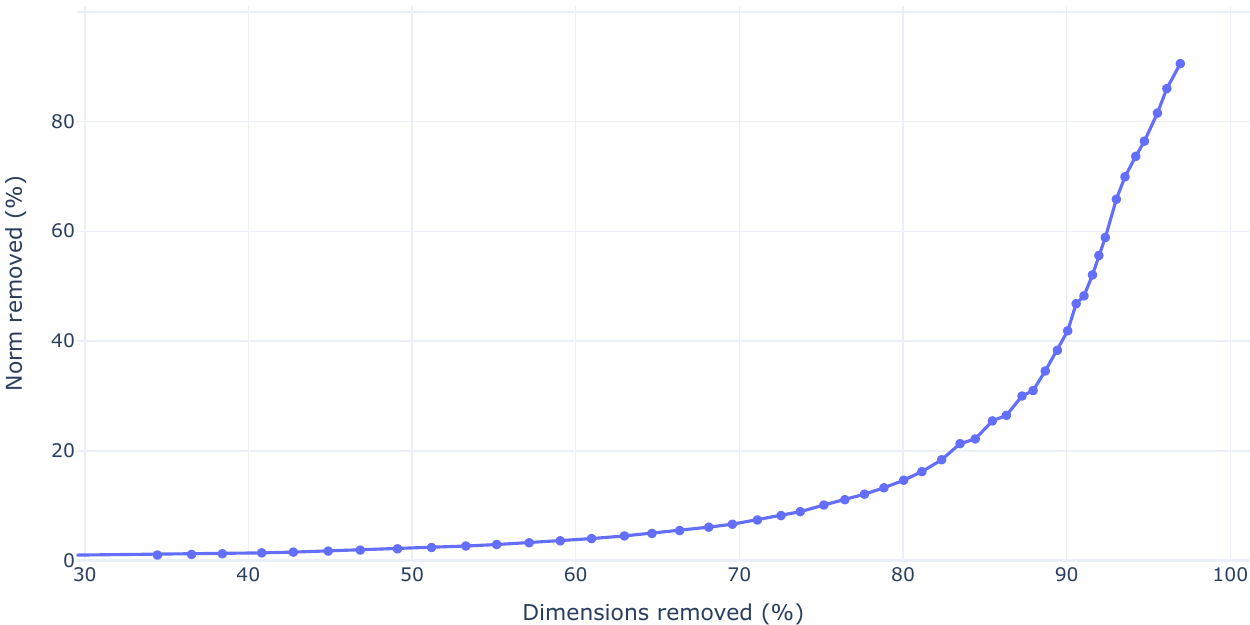}
    \caption{The tensor (Frobenius) norm as a function of removed dimensions. The norm is concentrated in only a fraction of the network.}
    \label{fig:trunc_norm}
\end{figure}

We measure which dimensions contribute to the Frobenius norm of the whole tensor (Figure~\ref{fig:trunc_norm}). This reveals only a handful of dimensions contribute.

\clearpage

\section{ODT in complexity} \label{app:complexity}
The overall time complexity of the ODT algorithm is $\mathcal{O}(L \cdot h^4)$, which is
linear in the depth $L$ of the network and quartic in the hidden dimension $h$. This section explains the complexity of the algorithm's steps per layer.

\subsubsection{Orthogonalisation} The complexity of a reduced RQ decomposition (for a matrix with dimensions $n$ and $m$) is $\mathcal{O}(\max(n, m) \cdot \min(n, m)^2)$. Each matricised core has dimensions $h$ and $h^2$, resulting in a complexity of $\mathcal{O}(h^4)$. Contracting the $R_i$ matrix into the next core $f_{i+1}$ (a third-order tensor) likewise has a complexity of $\mathcal{O}(h^4)$.

\subsubsection{Diagonalisation} Due to the imposed symmetry of the cores $f^\bot_i$, the projector $P_i$ only needs to be
calculated once per layer. Furthermore, one can avoid computing repeated tensor contractions by computing the Gram matrices $G_i$
iteratively from top to bottom using the isometric properties of the cores $f^\bot_i$. Consequently, each Gram matrix can be computed in $\mathcal{O}(h^4)$ time. The spectral decomposition of a Gram matrix can be computed in $\mathcal{O}(h^3)$ time, and contracting $V_i$ into their neighbouring cores takes $\mathcal{O}(h^4)$. 

\subsubsection{Truncation} If the eigenvalues $\Lambda_i$ from the previous step are stored, computing the truncation can be done in $\mathcal{O}(h)$ since it only involves removing dimensions. Contracting each projection into the adjacent cores can be done $\mathcal{O}(h^4)$.

In summary, the ODT algorithm has a bottom-up step (orthogonalisation) with complexity $\mathcal{O}(L \cdot h^4)$, a top-down step (diagonalisation) and a truncation step with the same complexity and a truncation step with negligible similar complexity. Despite the algorithm's quartic complexity, runtime rarely constrains performance. The $\mathcal{O}(h^3)$ memory requirement becomes the primary bottleneck for wide models.

\section{ODT in pictures and code} \label{app:diagrams}
Each of the three steps in the ODT algorithm discussed in Section~\ref{sec:ODT} is illustrated with string diagrammatic equations and pseudocode.

\subsubsection{Orthogonalisation} In Figure~\ref{fig:orthogonalisation} and Algorithm~\ref{alg:orthogonalisation}, cores $f_i$ (for $i=0, \dots, L$) are converted into isometries to achieve the isometric property shown in Figure~\ref{fig:isometric}, using the  elimination rule for the adjoint of the cloning operator:
\etz{cloning_adjoint}{0.8}

\subsubsection{Diagonalisation} The diagonalising projection matrices $P_i = V_i V_i^*$ for each bond $\mathcal{H}_i$ are constructed based on the eigenvectors of the gram matrix $G_i$, efficiently calculated as shown in Figure~\ref{fig:svd} and Algorithm~\ref{alg:diagonalisation}.

\subsubsection{Truncation} The truncated projections $P_i$ are contracted into the adjacent cores $f^\bot_{i-1}$ and $f^\bot_i$ in Figure~\ref{fig:truncation}. When all $P_i$ are contracted in parallel, a new $\rchi'$- net is formed as shown in Figure~\ref{fig:parallel} and Algorithm~\ref{alg:truncation}.

%%%%%%%%%%%%%%%%%%%%%%%%%%%%%%%%%%%%%%%%%%%%%%%%%%%%%%%%%%%%%%%

\begin{algorithm}
\caption{Orthogonalisation}
\label{alg:orthogonalisation}
\begin{algorithmic}[1]  % The [1] adds line numbers
    \REQUIRE Cores $f_0, \dots, f_{L + 1}$
    \STATE $f_0^{\not\bot} \leftarrow f_0$
    \FOR {$i=0$ to $L$}
        \STATE $R_i, Q_i \leftarrow \text{RQ}(f_i^{\not\bot})$
        \IF{$i = L$} 
            \STATE $f_{L+1}^{\not\bot} \leftarrow f_{L+1} \circ R_i$         \COMMENT{Unembedding}
        \ELSE
            \STATE $f_{i+1}^{\not\bot} \leftarrow f_{i+1} \circ (R_i \otimes R_i)$
        \ENDIF
        \STATE $f_i^\bot \leftarrow Q_i$
    \ENDFOR
    
    \RETURN $f_0^\bot, \dots, f_L^\bot, f_{L+1}^{\not\bot}$
\end{algorithmic}
\end{algorithm}
\begin{algorithm}
\caption{Diagonalisation}
\label{alg:diagonalisation}
\begin{algorithmic}[1]  % The [1] adds line numbers
    \REQUIRE Isometric cores $f^\bot_0, \dots, f^\bot_L$
    \REQUIRE Unembedding $f_{L + 1}^{\not\bot}$

    \FOR{$i = L + 1$ to $1$}
        \IF{$i = L + 1$}
            \STATE $G_{L + 1} \leftarrow f_{L + 1}^{\not\bot*} \circ f_{L + 1}^{\not\bot}$ \COMMENT{Unembedding}
        \ELSE
            \STATE $G_i \leftarrow f_i^{\bot*} \circ G_{i + 1} \circ f_i^{\bot}$
        \ENDIF

        \STATE $V_i, \Lambda_i \leftarrow \text{EVD}(G_i)$
    \ENDFOR

    \RETURN $V_1, \dots, V_{L + 1}, \Lambda_1, \dots, \Lambda_{L + 1}$
\end{algorithmic}
\end{algorithm}
\begin{algorithm}
\caption{Truncation}
\label{alg:truncation}
\begin{algorithmic}[1]  % The [1] adds line numbers
    \REQUIRE Isometric cores $f^\bot_0, \dots, f^\bot_L$
    \REQUIRE Unembedding $f_{L + 1}^{\not\bot}$
    \REQUIRE Isometries $V_1, \dots, V_{L + 1}$ of $r_1, \dots, r_{L + 1}$ most important singular vectors

    \STATE $f'_0 \leftarrow f_0^\bot$

    \FOR{$i = 1$ to $L + 1$}
        \STATE $f'_{i-1} \leftarrow V_i^* \circ f'_{i - 1}$
        \IF{$i = L + 1$}
            \STATE $f'_{L + 1} \leftarrow f_{L + 1}^{\not\bot} \circ V_{L + 1}$ \COMMENT{Unembedding}
        \ELSE
            \STATE $f'_{i} \leftarrow f_{i}^\bot \circ (V_i \otimes V_i)$
        \ENDIF
    \ENDFOR

    \RETURN $f'_0, \dots, f'_{L + 1}$
\end{algorithmic}
\end{algorithm}

%%%%%%%%%%%%%%%%%%%%%%%%%%%%%%%%%%%%%%%%%%%%%%%%%%%%%%%%%%%%%%%

\ctzwide{orthogonalisation}{0.7}{Orthogonalisation of a 3-layer $\rchi$-net by bottom-up application of the reduced RQ decomposition.}

\ctzwide{svd}{0.8}{Example diagonalisation of the Gram matrix $G_3$. Step 1 is a consequence of the isometric property of the cores, as shown in Figure~\ref{fig:isometric}. Step 2 is the spectral decomposition of the self-adjoint Gram matrix.}

\ctz{isometric}{0.8}{Isometric property of the orthogonalised $\rchi$-net. In the diagrammatic language, vertical reflection corresponds to taking the adjoint. The equality can be proven by contracting the isometries and the cloning operator using Equation~\ref{eq:cloning_adjoint}.}

\ctz{truncation}{0.8}{Example contraction of the truncated projector $P_3$ into the cores $f^\bot_2$ and $f^\bot_3$, resulting in a reduced bond dimension of $\mathcal{H}'_3$.}
\ctz{parallel}{0.8}{Example of a 3-layer $\rchi$-net after diagonalisation by the projectors $P_i$, shown on the left. If these $P_i$ are truncated to a lower rank, the resulting $\rchi'$-net is a low-rank approximation of the original $\rchi$-net, as shown on the right.}

%%%%%%%%%%%%%%%%%%%%%%%%%%%%%%%%%%%%%%%%%%%%%%%%%%%%%%%%%%%%%%%

\section*{Acknowledgements}
This research was funded by the Flemish Government under the "Onderzoeksprogramma Artificiële Intelligentie (AI) Vlaanderen” programme (TD, GAW, JO) and by the Department of Computer Science of the Vrije Universiteit Brussel (WG).

\section*{Contributions}
TD and WG contributed equally to the research and coauthored the manuscript. GW and JO guided both the research and writing process.

\bibliography{aaai25}

\end{document}

%% file: book.tikzstyles
\tikzstyle{single}=[draw=black,text=black]

\tikzstyle{env}=[copoint,regular polygon rotate=0,minimum width=0.2cm, fill=black]

\tikzstyle{probs}=[shape=semicircle,fill=white,draw=black,shape border rotate=180,minimum width=1.2cm]

\tikzstyle{diredges}=[every to/.style={diredge}]
\tikzstyle{math matrix}=[matrix of math nodes,left delimiter=(,right delimiter=),inner sep=2pt,column sep=1em,row sep=0.5em,nodes={inner sep=0pt},text height=1.5ex, text depth=0.25ex]

\tikzstyle{gs double edge}=[double,shorten <=-1mm,shorten >=-1mm,double distance=1pt]

\tikzstyle{inline text}=[text height=1.75ex, text depth=0.3ex,yshift=0mm]
\tikzstyle{label}=[font=\small,text height=1ex, text depth=0.15ex]
\tikzstyle{left label}=[label,anchor=east,xshift=1.5mm]
\tikzstyle{right label}=[label,anchor=west,xshift=-1.5mm]

\tikzstyle{braceedge}=[decorate,decoration={brace,amplitude=2mm,raise=-1mm}]
\tikzstyle{small braceedge}=[decorate,decoration={brace,amplitude=1mm,raise=-1mm}]
\tikzstyle{parenthesis}=[decorate,decoration={parenthesis,amplitude=2mm,raise=-1mm}]

\tikzstyle{doubled}=[line width=1.6pt] % set the line width for all doubled (quantum) maps/wires
\tikzstyle{boldedge}=[doubled,shorten <=-0.17mm,shorten >=-0.17mm]
\tikzstyle{boldedgegray}=[doubled,gray,shorten <=-0.17mm,shorten >=-0.17mm]
\tikzstyle{singleedgegray}=[gray]%,shorten <=-0.1mm,shorten >=-0.1mm]

\tikzstyle{semidoubled}=[line width=1.4pt] % set the line width for all doubled (quantum) maps/wires
\tikzstyle{semiboldedgegray}=[semidoubled,gray,shorten <=-0.17mm,shorten >=-0.17mm]

\tikzstyle{boxedge}=[semiboldedgegray]

\tikzstyle{boldedgedashed}=[very thick,dashed,shorten <=-0.17mm,shorten >=-0.17mm]
\tikzstyle{vboldedgedashed}=[doubled,dashed,shorten <=-0.17mm,shorten >=-0.17mm]
\tikzstyle{left hook arrow}=[left hook-latex]
\tikzstyle{right hook arrow}=[right hook-latex]
\tikzstyle{sembracket}=[line width=0.5pt,shorten <=-0.07mm,shorten >=-0.07mm]

\tikzstyle{causal edge}=[->,thick,gray]
\tikzstyle{causal nondir}=[thick,gray]
\tikzstyle{timeline}=[thick,gray, dashed]

\tikzstyle{cedge}=[<->,thick,gray!70!white]

\tikzstyle{empty diagram}=[draw=gray!40!white,dashed,shape=rectangle,minimum width=1cm,minimum height=1cm]
\tikzstyle{empty diagram small}=[draw=gray!50!white,dashed,shape=rectangle,minimum width=0.6cm,minimum height=0.5cm]

\tikzstyle{dot}=[inner sep=0mm,minimum width=2mm,minimum height=2mm,draw,shape=circle]  
\tikzstyle{Wsquare}=[white dot, shape=regular polygon, rounded corners=0.8 mm, minimum size=3.3 mm, regular polygon sides=3, outer sep=-0.2mm]
\tikzstyle{Wsquareadj}=[white dot, shape=regular polygon, rounded corners=0.8 mm, minimum size=3.3 mm, regular polygon sides=3, outer sep=-0.2mm, regular polygon rotate=180]
\tikzstyle{ddot}=[inner sep=0mm, doubled, minimum width=2.5mm,minimum height=2.5mm,draw,shape=circle]

\tikzstyle{black dot}=[dot,fill=black]
\tikzstyle{white dot}=[dot,fill=white,text depth=-0.2mm]
\tikzstyle{white Wsquare}=[Wsquare,fill=white,text depth=-0.2mm]
\tikzstyle{white Wsquareadj}=[Wsquareadj,fill=white,text depth=-0.2mm]
\tikzstyle{green dot}=[white dot] % for backwards-compatibility
\tikzstyle{gray dot}=[dot,fill=gray!40!white,text depth=-0.2mm]
\tikzstyle{red dot}=[gray dot] % for backwards-compatibility

\tikzstyle{Z}=[white dot]
\tikzstyle{X}=[gray dot]
\tikzstyle{black ddot}=[ddot,fill=black]
\tikzstyle{white ddot}=[ddot,fill=white]
\tikzstyle{gray ddot}=[ddot,fill=gray!40!white]

\tikzstyle{gray edge}=[gray!60!white]

\tikzstyle{small dot}=[inner sep=0.5mm,minimum width=0pt,minimum height=0pt,draw,shape=circle]

\tikzstyle{small black dot}=[small dot,fill=black]
\tikzstyle{small white dot}=[small dot,fill=white]
\tikzstyle{small gray dot}=[small dot,fill=gray!40!white]

\tikzstyle{mbqc dot}=[small black dot]
\tikzstyle{mbqc input dot}=[small white dot]
\tikzstyle{mbqc output dot}=[small gray dot]

\tikzstyle{causal dot}=[inner sep=0.4mm,minimum width=0pt,minimum height=0pt,draw=white,shape=circle,fill=gray!40!white]

\tikzstyle{phase dimensions}=[minimum size=5mm,font=\footnotesize,rectangle,rounded corners=2mm,inner sep=0.2mm,outer sep=-2mm,scale=0.8]
\tikzstyle{dphase dimensions}=[minimum size=5mm,font=\footnotesize,rectangle,rounded corners=2.5mm,inner sep=0.2mm,outer sep=-2mm]
\tikzstyle{white phase dot}=[dot,fill=white,phase dimensions]
\tikzstyle{white phase ddot}=[ddot,fill=white,dphase dimensions]

\tikzstyle{white rect ddot}=[draw=black,fill=white,doubled,minimum size=5mm,font=\footnotesize,rectangle,rounded corners=2.5mm,inner sep=0.2mm]
\tikzstyle{gray rect ddot}=[draw=black,fill=gray!40!white,doubled,minimum size=6mm,font=\footnotesize,rectangle,rounded corners=3mm]

\tikzstyle{gray phase dot}=[dot,fill=gray!40!white,phase dimensions]
\tikzstyle{gray phase ddot}=[ddot,fill=gray!40!white,dphase dimensions]
\tikzstyle{grey phase dot}=[gray phase dot]
\tikzstyle{grey phase ddot}=[gray phase ddot]

\tikzstyle{small phase dimensions}=[minimum size=4mm,font=\tiny,rectangle,rounded corners=2mm,inner sep=0.2mm,outer sep=-2mm]
\tikzstyle{small dphase dimensions}=[minimum size=4mm,font=\tiny,rectangle,rounded corners=2mm,inner sep=0.2mm,outer sep=-2mm]

\tikzstyle{small gray phase dot}=[dot,fill=gray!40!white,small phase dimensions]
\tikzstyle{small gray phase ddot}=[ddot,fill=gray!40!white,small dphase dimensions]

\tikzstyle{small map}=[draw,shape=rectangle,minimum height=4mm,minimum width=4mm,fill=white]

\tikzstyle{cnot}=[fill=white,shape=circle,inner sep=-1.4pt]

\tikzstyle{asym hadamard}=[fill=white,draw,shape=NEbox,inner sep=0.6mm,font=\footnotesize,minimum height=4mm]
\tikzstyle{asym hadamard conj}=[fill=white,draw,shape=NWbox,inner sep=0.6mm,font=\footnotesize,minimum height=4mm]
\tikzstyle{asym hadamard dag}=[fill=white,draw,shape=SEbox,inner sep=0.6mm,font=\footnotesize,minimum height=4mm]

\tikzstyle{clone}=[
    fill=black,
    shape=isosceles triangle,
    isosceles triangle apex angle=60,
    draw,
    rotate=90,
    inner sep=0.6mm,
    font=\footnotesize
    ]
\tikzstyle{clone dag}=[clone,rotate=180]

\tikzstyle{hadamard}=[fill=white,draw,inner sep=0.6mm,font=\footnotesize,minimum height=4mm,minimum width=4mm]
\tikzstyle{small hadamard}=[fill=white,draw,inner sep=0.6mm,minimum height=1.5mm,minimum width=1.5mm]
\tikzstyle{small hadamard rotate}=[small hadamard,rotate=45]
\tikzstyle{dhadamard}=[hadamard,doubled]
\tikzstyle{small dhadamard}=[small hadamard,doubled]
\tikzstyle{small dhadamard rotate}=[small hadamard rotate,doubled]
\tikzstyle{antipode}=[white dot,inner sep=0.3mm,font=\footnotesize]

\tikzstyle{scalar}=[diamond,draw,inner sep=0.5pt,font=\small]
\tikzstyle{dscalar}=[diamond,doubled, draw,inner sep=0.5pt,font=\small]

\tikzstyle{small box}=[rectangle,inline text,fill=white,draw,minimum height=5mm,yshift=-0.5mm,minimum width=5mm,font=\small]
\tikzstyle{small gray box}=[small box,fill=gray!30]
\tikzstyle{medium box}=[rectangle,inline text,fill=white,draw,minimum height=5mm,yshift=-0.5mm,minimum width=10mm,font=\small]
\tikzstyle{square box}=[small box] % for backwards-compatibility
\tikzstyle{medium gray box}=[small box,fill=gray!30]
\tikzstyle{semilarge box}=[rectangle,inline text,fill=white,draw,minimum height=5mm,yshift=-0.5mm,minimum width=12.5mm,font=\small]
\tikzstyle{large box}=[rectangle,inline text,fill=white,draw,minimum height=5mm,yshift=-0.5mm,minimum width=15mm,font=\small]
\tikzstyle{large gray box}=[small box,fill=gray!30]

\tikzstyle{Bayes box}=[rectangle,fill=black,draw, minimum height=3mm, minimum width=3mm]

\tikzstyle{gray square point}=[small box,fill=gray!50]

\tikzstyle{dphase box white}=[dhadamard]
\tikzstyle{dphase box gray}=[dhadamard,fill=gray!50!white]
\tikzstyle{phase box white}=[hadamard]
\tikzstyle{phase box gray}=[hadamard,fill=gray!50!white]

\tikzstyle{point}=[regular polygon,regular polygon sides=3,draw,scale=0.75,inner sep=-0.5pt,minimum width=9mm,fill=white,regular polygon rotate=180]
\tikzstyle{copoint}=[regular polygon,regular polygon sides=3,draw,scale=0.75,inner sep=-0.5pt,minimum width=9mm,fill=white]
\tikzstyle{dpoint}=[point,doubled]
\tikzstyle{dcopoint}=[copoint,doubled]

\tikzstyle{wide copoint}=[fill=white,draw,shape=isosceles triangle,shape border rotate=90,isosceles triangle stretches=true,inner sep=0pt,minimum width=1.5cm,minimum height=6.12mm]
\tikzstyle{wide point}=[fill=white,draw,shape=isosceles triangle,shape border rotate=-90,isosceles triangle stretches=true,inner sep=0pt,minimum width=1.5cm,minimum height=6.12mm,yshift=-0.0mm]
\tikzstyle{wide point plus}=[fill=white,draw,shape=isosceles triangle,shape border rotate=-90,isosceles triangle stretches=true,inner sep=0pt,minimum width=1.74cm,minimum height=7mm,yshift=-0.0mm]

\tikzstyle{wide dpoint}=[fill=white,doubled,draw,shape=isosceles triangle,shape border rotate=-90,isosceles triangle stretches=true,inner sep=0pt,minimum width=1.5cm,minimum height=6.12mm,yshift=-0.0mm]

\tikzstyle{tinypoint}=[regular polygon,regular polygon sides=3,draw,scale=0.55,inner sep=-0.15pt,minimum width=6mm,fill=white,regular polygon rotate=180] 

\tikzstyle{white point}=[point]
\tikzstyle{white dpoint}=[dpoint]
\tikzstyle{green point}=[white point] % for backwards-compatibility
\tikzstyle{white copoint}=[copoint]
\tikzstyle{gray point}=[point,fill=gray!40!white]
\tikzstyle{gray dpoint}=[gray point,doubled]
\tikzstyle{red point}=[gray point] % for backwards-compatibility
\tikzstyle{gray copoint}=[copoint,fill=gray!40!white]
\tikzstyle{gray dcopoint}=[gray copoint,doubled]

\tikzstyle{white point guide}=[regular polygon,regular polygon sides=3,font=\scriptsize,draw,scale=0.65,inner sep=-0.5pt,minimum width=9mm,fill=white,regular polygon rotate=180]

\tikzstyle{tiny gray point}=[tinypoint,fill=gray!40!white]

\tikzstyle{diredge}=[->]
\tikzstyle{ddiredge}=[<->]
\tikzstyle{rdiredge}=[<-]
\tikzstyle{thickdiredge}=[->, very thick]
\tikzstyle{pointer edge}=[->,very thick,gray]
\tikzstyle{pointer edge part}=[very thick,gray]
\tikzstyle{dashed edge}=[dashed]
\tikzstyle{thick dashed edge}=[very thick,dashed]
\tikzstyle{thick gray dashed edge}=[thick dashed edge,gray!40]
\tikzstyle{gray dashed edge}=[dashed edge,gray!50]
\tikzstyle{thick map edge}=[very thick,|->]

\tikzstyle{cloud}=[shape=cloud,draw,minimum width=1.5cm,minimum height=1.5cm]

\tikzstyle{map}=[draw,shape=NEbox,inner sep=2pt,minimum height=6mm,fill=white]
\tikzstyle{dashedmap}=[draw,dashed,shape=NEbox,inner sep=2pt,minimum height=6mm,fill=white]
\tikzstyle{mapdag}=[draw,shape=SEbox,inner sep=2pt,minimum height=6mm,fill=white]
\tikzstyle{mapadj}=[draw,shape=SEbox,inner sep=2pt,minimum height=6mm,fill=white]
\tikzstyle{maptrans}=[draw,shape=SWbox,inner sep=2pt,minimum height=6mm,fill=white]
\tikzstyle{mapconj}=[draw,shape=NWbox,inner sep=2pt,minimum height=6mm,fill=white]

\tikzstyle{medium map}=[draw,shape=NEbox,inner sep=2pt,minimum height=6mm,fill=white,minimum width=7mm]
\tikzstyle{medium map dag}=[draw,shape=SEbox,inner sep=2pt,minimum height=6mm,fill=white,minimum width=7mm]
\tikzstyle{medium map adj}=[draw,shape=SEbox,inner sep=2pt,minimum height=6mm,fill=white,minimum width=7mm]
\tikzstyle{medium map trans}=[draw,shape=SWbox,inner sep=2pt,minimum height=6mm,fill=white,minimum width=7mm]
\tikzstyle{medium map conj}=[draw,shape=NWbox,inner sep=2pt,minimum height=6mm,fill=white,minimum width=7mm]
\tikzstyle{semilarge map}=[draw,shape=NEbox,inner sep=2pt,minimum height=6mm,fill=white,minimum width=9.5mm]
\tikzstyle{semilarge map trans}=[draw,shape=SWbox,inner sep=2pt,minimum height=6mm,fill=white,minimum width=9.5mm]
\tikzstyle{semilarge map adj}=[draw,shape=SEbox,inner sep=2pt,minimum height=6mm,fill=white,minimum width=9.5mm]
\tikzstyle{semilarge map dag}=[draw,shape=SEbox,inner sep=2pt,minimum height=6mm,fill=white,minimum width=9.5mm]
\tikzstyle{semilarge map conj}=[draw,shape=NWbox,inner sep=2pt,minimum height=6mm,fill=white,minimum width=9.5mm]
\tikzstyle{large map}=[draw,shape=NEbox,inner sep=2pt,minimum height=6mm,fill=white,minimum width=12mm]
\tikzstyle{large map dag}=[draw,shape=SEbox,inner sep=2pt,minimum height=6mm,fill=white,minimum width=12mm]
\tikzstyle{large map conj}=[draw,shape=NWbox,inner sep=2pt,minimum height=6mm,fill=white,minimum width=12mm]
\tikzstyle{very large map}=[draw,shape=NEbox,inner sep=2pt,minimum height=6mm,fill=white,minimum width=17mm]

\tikzstyle{medium dmap}=[draw,doubled,shape=NEbox,inner sep=2pt,minimum height=6mm,fill=white,minimum width=7mm]
\tikzstyle{medium dmap dag}=[draw,doubled,shape=SEbox,inner sep=2pt,minimum height=6mm,fill=white,minimum width=7mm]
\tikzstyle{medium dmap adj}=[draw,doubled,shape=SEbox,inner sep=2pt,minimum height=6mm,fill=white,minimum width=7mm]
\tikzstyle{medium dmap trans}=[draw,doubled,shape=SWbox,inner sep=2pt,minimum height=6mm,fill=white,minimum width=7mm]
\tikzstyle{medium dmap conj}=[draw,doubled,shape=NWbox,inner sep=2pt,minimum height=6mm,fill=white,minimum width=7mm]
\tikzstyle{semilarge dmap}=[draw,doubled,shape=NEbox,inner sep=2pt,minimum height=6mm,fill=white,minimum width=9.5mm]
\tikzstyle{semilarge dmap trans}=[draw,doubled,shape=SWbox,inner sep=2pt,minimum height=6mm,fill=white,minimum width=9.5mm]
\tikzstyle{semilarge dmap adj}=[draw,doubled,shape=SEbox,inner sep=2pt,minimum height=6mm,fill=white,minimum width=9.5mm]
\tikzstyle{semilarge dmap dag}=[draw,doubled,shape=SEbox,inner sep=2pt,minimum height=6mm,fill=white,minimum width=9.5mm]
\tikzstyle{semilarge dmap conj}=[draw,doubled,shape=NWbox,inner sep=2pt,minimum height=6mm,fill=white,minimum width=9.5mm]
\tikzstyle{large dmap}=[draw,doubled,shape=NEbox,inner sep=2pt,minimum height=6mm,fill=white,minimum width=12mm]
\tikzstyle{large dmap conj}=[draw,doubled,shape=NWbox,inner sep=2pt,minimum height=6mm,fill=white,minimum width=12mm]
\tikzstyle{large dmap trans}=[draw,doubled,shape=SWbox,inner sep=2pt,minimum height=6mm,fill=white,minimum width=12mm]
\tikzstyle{large dmap adj}=[draw,doubled,shape=SEbox,inner sep=2pt,minimum height=6mm,fill=white,minimum width=12mm]
\tikzstyle{large dmap dag}=[draw,doubled,shape=SEbox,inner sep=2pt,minimum height=6mm,fill=white,minimum width=12mm]
\tikzstyle{very large dmap}=[draw,doubled,shape=NEbox,inner sep=2pt,minimum height=6mm,fill=white,minimum width=19.5mm]
\tikzstyle{huge dmap}=[draw,doubled,shape=SEbox,inner sep=2pt,minimum height=6mm,fill=white,minimum width=30mm]

\tikzstyle{muxbox}=[draw,shape=rectangle,minimum height=3mm,minimum width=3mm,fill=white]
\tikzstyle{dmuxbox}=[muxbox,doubled]

\tikzstyle{box}=[draw,shape=rectangle,inner sep=2pt,minimum height=6mm,minimum width=6mm,fill=white]
\tikzstyle{dbox}=[draw,doubled,shape=rectangle,inner sep=2pt,minimum height=6mm,minimum width=6mm,inline text,fill=white]
\tikzstyle{dmap}=[draw,doubled,shape=NEbox,inner sep=2pt,minimum height=6mm,fill=white]
\tikzstyle{dmapdag}=[draw,doubled,shape=SEbox,inner sep=2pt,minimum height=6mm,fill=white]
\tikzstyle{dmapadj}=[draw,doubled,shape=SEbox,inner sep=2pt,minimum height=6mm,fill=white]
\tikzstyle{dmaptrans}=[draw,doubled,shape=SWbox,inner sep=2pt,minimum height=6mm,fill=white]
\tikzstyle{dmapconj}=[draw,doubled,shape=NWbox,inner sep=2pt,minimum height=6mm,fill=white]

\tikzstyle{ddmap}=[draw,doubled,dashed,shape=NEbox,inner sep=2pt,minimum height=6mm,fill=white]
\tikzstyle{ddmapdag}=[draw,doubled,dashed,shape=SEbox,inner sep=2pt,minimum height=6mm,fill=white]
\tikzstyle{ddmapadj}=[draw,doubled,dashed,shape=SEbox,inner sep=2pt,minimum height=6mm,fill=white]
\tikzstyle{ddmaptrans}=[draw,doubled,dashed,shape=SWbox,inner sep=2pt,minimum height=6mm,fill=white]
\tikzstyle{ddmapconj}=[draw,doubled,dashed,shape=NWbox,inner sep=2pt,minimum height=6mm,fill=white]

\tikzstyle{smap}=[draw,shape=sNEbox,fill=white]
\tikzstyle{smapdag}=[draw,shape=sSEbox,fill=white]
\tikzstyle{smapadj}=[draw,shape=sSEbox,fill=white]
\tikzstyle{smaptrans}=[draw,shape=sSWbox,fill=white]
\tikzstyle{smapconj}=[draw,shape=sNWbox,fill=white]

\tikzstyle{dsmap}=[draw,dashed,shape=sNEbox,fill=white]
\tikzstyle{dsmapdag}=[draw,dashed,shape=sSEbox,fill=white]
\tikzstyle{dsmaptrans}=[draw,dashed,shape=sSWbox,fill=white]
\tikzstyle{dsmapconj}=[draw,dashed,shape=sNWbox,fill=white]

\tikzstyle{mmap}=[draw,shape=mNEbox]
\tikzstyle{mmapdag}=[draw,shape=mSEbox]
\tikzstyle{mmaptrans}=[draw,shape=mSWbox]
\tikzstyle{mmapconj}=[draw,shape=mNWbox]

\tikzstyle{mmapgray}=[draw,fill=gray!40!white,shape=mNEbox]
\tikzstyle{smapgray}=[draw,fill=gray!40!white,shape=sNEbox]

\tikzstyle{kpoint common}=[draw,fill=white,inner sep=1pt,minimum height=4mm]
\tikzstyle{kpoint sc}=[shape=cornerpoint,kpoint common]
\tikzstyle{kpoint adjoint sc}=[shape=cornercopoint,kpoint common]
\tikzstyle{kpoint}=[shape=cornerpoint,shorten left=5pt,kpoint common]
\tikzstyle{kpoint adjoint}=[shape=cornercopoint,shorten left=5pt,kpoint common]
\tikzstyle{kpoint conjugate}=[shape=cornerpoint,shorten right=5pt,kpoint common]
\tikzstyle{kpoint transpose}=[shape=cornercopoint,shorten right=5pt,kpoint common]
\tikzstyle{kpoint symm}=[shape=cornerpoint,shorten left=5pt,shorten right=5pt,kpoint common]

\tikzstyle{kpointdag}=[kpoint adjoint]
\tikzstyle{kpointadj}=[kpoint adjoint]
\tikzstyle{kpointconj}=[kpoint conjugate]
\tikzstyle{kpointtrans}=[kpoint transpose]

\tikzstyle{big kpoint}=[kpoint, minimum width=1.2 cm, minimum height=8mm, inner sep=4pt, text depth=3mm]

\tikzstyle{wide kpoint}=[kpoint, minimum width=1 cm, inner sep=2pt]%, text depth=-0.7 mm]
\tikzstyle{wide kpointdag}=[kpointdag, minimum width=1 cm, inner sep=2pt]%, text depth=0.7 mm]
\tikzstyle{wide kpointconj}=[kpointconj, minimum width=1 cm, inner sep=2pt]%, text depth=-0.7 mm]
\tikzstyle{wide kpointtrans}=[kpointtrans, minimum width=1 cm, inner sep=2pt]%, text depth=0.7 mm]

\tikzstyle{gray kpoint}=[kpoint,fill=gray!50!white]
\tikzstyle{gray kpointdag}=[kpointdag,fill=gray!50!white]
\tikzstyle{gray kpointadj}=[kpointadj,fill=gray!50!white]
\tikzstyle{gray kpointconj}=[kpointconj,fill=gray!50!white]
\tikzstyle{gray kpointtrans}=[kpointtrans,fill=gray!50!white]

\tikzstyle{gray dkpoint}=[kpoint,fill=gray!50!white,doubled]
\tikzstyle{gray dkpointdag}=[kpointdag,fill=gray!50!white,doubled]
\tikzstyle{gray dkpointadj}=[kpointadj,fill=gray!50!white,doubled]
\tikzstyle{gray dkpointconj}=[kpointconj,fill=gray!50!white,doubled]
\tikzstyle{gray dkpointtrans}=[kpointtrans,fill=gray!50!white,doubled]

\tikzstyle{white label}=[draw,fill=white,rectangle,inner sep=0.7 mm]
\tikzstyle{gray label}=[draw,fill=gray!50!white,rectangle,inner sep=0.7 mm]
\tikzstyle{black label}=[draw,fill=black,rectangle,inner sep=0.7 mm]

\tikzstyle{dkpoint}=[kpoint,doubled]
\tikzstyle{wide dkpoint}=[wide kpoint,doubled]
\tikzstyle{dkpointdag}=[kpoint adjoint,doubled]
\tikzstyle{wide dkpointdag}=[wide kpointdag,doubled]
\tikzstyle{dkcopoint}=[kpoint adjoint,doubled]
\tikzstyle{dkpointadj}=[kpoint adjoint,doubled]
\tikzstyle{dkpointconj}=[kpoint conjugate,doubled]
\tikzstyle{dkpointtrans}=[kpoint transpose,doubled]

\tikzstyle{kscalar}=[kpoint common, shape=EBox, inner xsep=-1pt, inner ysep=3pt,font=\small]
\tikzstyle{kscalarconj}=[kpoint common, shape=WBox, inner xsep=-1pt, inner ysep=3pt,font=\small]

\tikzstyle{spekpoint}=[kpoint sc,minimum height=5mm,inner sep=3pt]
\tikzstyle{spekcopoint}=[kpoint adjoint sc,minimum height=5mm,inner sep=3pt]

\tikzstyle{dspekpoint}=[spekpoint,doubled]
\tikzstyle{dspekcopoint}=[spekcopoint,doubled]

 \tikzstyle{upground}=[circuit ee IEC,thick,ground,rotate=90,scale=2.7]
 \tikzstyle{downground}=[circuit ee IEC,thick,ground,rotate=-90,scale=2.5]
\tikzstyle{bigground}=[regular polygon,regular polygon sides=3,draw=gray,scale=0.50,inner sep=-0.5pt,minimum width=10mm,fill=gray]
\tikzstyle{arrs}=[-latex,font=\small,auto]
\tikzstyle{arrow plain}=[arrs]
\tikzstyle{arrow dashed}=[dashed,arrs]
\tikzstyle{arrow bold}=[very thick,arrs]
\tikzstyle{arrow hide}=[draw=white!0,-]
\tikzstyle{arrow reverse}=[latex-]